\documentclass[10pt,twocolumn,letterpaper]{article}

\usepackage{wacv}              
\usepackage{makecell}

\usepackage{graphicx}
\usepackage{amsmath}
\usepackage{amssymb}
\usepackage{booktabs}
\usepackage{algorithm}
\usepackage{algpseudocode}
\usepackage{enumitem}
\usepackage{tikz}
\usetikzlibrary{arrows.meta, positioning, shapes, matrix}
\usepackage{float}
\usepackage{booktabs}
\usepackage{multirow}
\usepackage{bbm}
\usepackage[pagebackref,breaklinks,colorlinks]{hyperref}
\usepackage[capitalize]{cleveref}
\crefname{section}{Sec.}{Secs.}
\Crefname{section}{Section}{Sections}
\Crefname{table}{Table}{Tables}
\crefname{table}{Tab.}{Tabs.}

\def\wacvPaperID{1066} 
\def\confName{WACV}
\def\confYear{2026}

\newcommand{\namex}{Polymorph}

\title{\namex: Energy-Efficient Multi-Label Classification for Video Streams on Embedded Devices}

\author{
Saeid Ghafouri$^{1}$ \quad
Mohsen Fayyaz$^{2}$ \quad
Xiangchen Li$^{3}$ \quad
Deepu John$^{4}$ \\
Bo Ji$^{3}$ \quad
Dimitrios Nikolopoulos$^{3}$ \quad
Hans Vandierendonck$^{1}$ \\
\\
$^{1}$Queen’s University Belfast, United Kingdom \\
$^{2}$Microsoft, Berlin, Germany \\
$^{3}$Virginia Tech, Blacksburg, VA, USA \\
$^{4}$University College Dublin, Ireland
}

\begin{document}

\maketitle

\begin{abstract}
Real-time multi-label video classification on embedded devices is constrained by limited compute and energy budgets. Yet, video streams exhibit structural properties such as label sparsity, temporal continuity, and label co-occurrence that can be leveraged for more efficient inference. We introduce \textbf{Polymorph}, a context-aware framework that activates a minimal set of lightweight Low Rank Adapters (LoRA) per frame. Each adapter specializes in a subset of classes derived from co-occurrence patterns and is implemented as a LoRA weight over a shared backbone. At runtime, Polymorph dynamically selects and composes only the adapters needed to cover the active labels, avoiding full-model switching and weight merging. This modular strategy improves scalability while reducing latency, and energy overhead. Polymorph achieves 40\% lower energy consumption and improves mAP by 9 points over strong baselines executing the TAO dataset. \namex~is open source at \url{https://github.com/inference-serving/polymorph/}.
\end{abstract}

\section{Introduction}

Deep learning has enabled major advances in visual recognition tasks, but running these models efficiently on embedded devices remains a core challenge. Applications such as smart surveillance cameras~\cite{ke2020surveillance}, mobile robotics~\cite{groshev2023robotics}, and wearable health monitors~\cite{incel2023health} increasingly demand real-time classification with strict constraints on compute, memory, and energy. These deployment scenarios rule out the use of large-scale, high-capacity models in their original form. A wide range of model compression techniques, such as pruning \cite{cheng2024pruning}, quantization~\cite{rokh2023quantization}, and architecture search~\cite{ren2021nas}, have made progress in reducing the size and latency of deep models. However, these approaches tend to plateau in terms of accuracy when pushed to the extreme resource limits required by on-device deployment. As a result, there is a growing need for new paradigms that can maintain high performance while operating within the tight energy budgets of embedded systems.

Among these diverse application of deep learning, multi-label classification has been widely studied in the context of image recognition, where models must predict multiple labels that may co-occur within a single input~\cite{Chen_2019_CVPR,Zhao_2021_ICCV,Xia_2023_ICCV}. 
In particular, multi-label classification over video streams presents a unique set of challenges. Unlike static image classification, video inputs are continuous and evolving, with different objects and semantic labels appearing and disappearing across time. Each frame may contain multiple labels, and the set of active labels can vary significantly between frames. Importantly, the number of potential labels is often large, but the number of labels per frame is typically small (up to 3). This combination creates tension between accuracy and efficiency: large models are needed to have reasonable accuracy for a large set of classes, but only a few number of classes are relevant at any given moment. Our goal in this work is to design a system that reduces energy on embedded devices for multi-label classification on video streams, while maintaining or improving predictive accuracy. To achieve this, we identified three structural properties of mutlti-label classification task of video stream:

\begin{itemize}
    \item \textbf{Label sparsity:} Only a small subset of classes appears in each frame.
    \item \textbf{Temporal continuity:} The set of active labels often changes gradually over time.
    \item \textbf{Label co-occurrence structure:} Certain classes frequently appear together, forming natural groupings.
\end{itemize}

These observations suggest that it is unnecessary to rely on a single large model to perform multi-label classification across all video frames. Instead, computation can be adaptively focused on a smaller subset of classes, which we refer to as the video \textit{context}. A context is defined as a group of labels that frequently co-occur within a short temporal segment. A context change occurs when the active label set in a frame differs from previous frames, typically due to new object types appearing or others disappearing.

\begin{figure}[t]
    \includegraphics[width=0.9\linewidth]{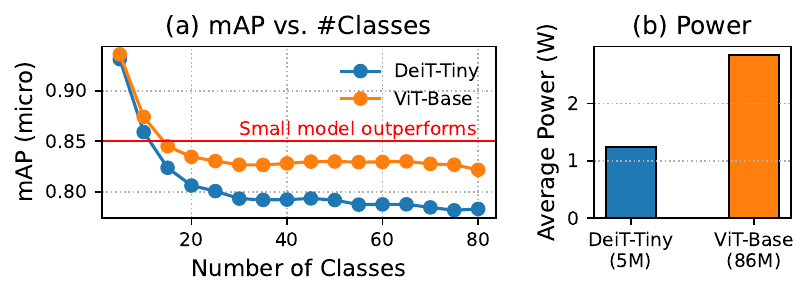}
    \caption{(a) Smaller model can surpass larger variant accuracy on smaller subset of classes (5-80 increments of 5), comparing two ViT variants trained on COCO dataset for multi-label classification task, (b) ViT base and DeiT tiny with 86.6 M and 5 M parameters count
    \label{fig:larger-smaler-models} and 2x less energy consumption.}
    \vspace{-1em}
\end{figure}

Training on fewer, semantically related classes can improve accuracy, as the model does not need to allocate capacity to unrelated or absent labels~\cite{rastikerdar2024cactus, hu2022lora}. As illustrated in Figure~\ref{fig:larger-smaler-models}~(a), a DeiT-tiny model (5M parameters)~\cite{dosovitskiy2021an} trained on a reduced class set (e.g., five classes) achieves higher mAP (mean Average Precision) than a ViT-base model (86.6M parameters)~\cite{touvron2021training} trained on the other label set sizes (e.g. 20, 40, 60 and 80 classes), while consuming half the energy (Figure~\ref{fig:larger-smaler-models}~(b)).

We build on this idea to propose \textbf{Polymorph}, a context-aware framework that dynamically selects and combines specialized classifiers at runtime. Each classifier is trained on a subset of labels derived from co-occurrence patterns in the dataset, capturing the natural structure of label groupings. To make this approach efficient and composable, we implement each classifier as a lightweight \emph{LoRA} (Low-Rank Adaptation) \cite{hu2022lora} adapter. LoRA is a technique that injects trainable, low-rank parameter matrices into a frozen backbone model, enabling fast adaptation with minimal additional parameters. In our setting, this allows each context-specific classifier to be expressed as a small LoRA adapter over a shared backbone, avoiding the need to store or switch between full models. To support this sharing, we apply LoRA only to the final layers of the network, keeping the rest of the backbone frozen and common across all contexts.

Unlike prior context-aware approaches such as CACTUS~\cite{rastikerdar2024cactus}, which assume only one active model at a time, Polymorph allows multiple LoRA adapters to be active simultaneously. This supports flexible, efficient inference even when the active labels in a frame span multiple contexts. Furthermore, because LoRA adapters are small and only adapt a portion of the model’s weights, we avoid the memory and switching costs of full-model approaches. Our system selects a minimal set of LoRA adapters needed to cover the frame’s labels. To effectively assign LoRAs to each class, we designed a two-stage context selection strategy. At training time, we group classes into context sets using clustering over their co-occurrence patterns, ensuring that each LoRA adapter captures a semantically meaningful subset of labels. At inference time, we select the minimal subset of these LoRA adapters needed to cover the active labels in the current frame.

In summary, we propose \textbf{\namex}, a context-aware and LoRA-based framework for efficient multi-label classification in video streams. Our main contributions are:

\begin{itemize}
    \item We identify and exploit key structural properties of video streams, namely, label sparsity, temporal continuity, and co-occurrence structure, to reduce inference energy cost without sacrificing accuracy.

    \item We propose a two-stage context selection strategy that uses label co-occurrence clustering at training time and a runtime context detection at inference time to activate only the relevant LoRA adapters per frame.

    \item We design a composable inference mechanism where each context is a lightweight LoRA adapter applied only to the final layers of a shared backbone. This allows most computation to be shared, enabling parallel execution of multiple adapters with low latency and energy overhead.

    \item We empirically demonstrate that \namex\ achieves substantial improvements in energy efficiency by 40\% compared to all-class models and single-classifier switching baselines, while  improving mAP by 9.
\end{itemize}

\section{Related Work}
\label{sec:related}

We categorize highly related works into four groups.

\textbf{Quantization and Pruning for Edge Deployment.} These techniques reduce model size and computation for real-time inference. \textit{Edge-MPQ}~\cite{zhao2024edge} applies layer-wise mixed precision on a co-designed hardware-software stack. \textit{PiQi}~\cite{aghapour2024piqi} integrates partial quantization and model partitioning. \textit{Agile-Quant}~\cite{shen2024agile} adapts precision to activation sensitivity. \textit{8-bit Transformer}~\cite{yu20248} enables full FP8/Posit8 inference and fine-tuning with BFloat16-level accuracy. TensorRT and ONNX Runtime~\cite{nvidia_tensorrt_2025, onnx_2025} support efficient deployment across hardware. These methods can be applied within each LoRA adapter or main model orthogonal to our approach.

\textbf{Mixture-of-Experts for Efficient Inference.} Mixture-of-Experts (MoE) models improve efficiency by activating only a subset of components per input, reducing computation while maintaining capacity. General MoE designs like GShard~\cite{lepikhingshard} and Switch Transformer~\cite{fedus2022switch} scale via sparse activation and simple routing. For edge use, \textit{Mobile V-MoEs}~\cite{daxberger2023mobile} adapt ViTs with sparse MoE layers and semantic routing for constrained devices, while \textit{Edge-MoE}~\cite{sarkar2023edge} offers a memory-efficient multi-task ViT with task-level sparsity.

In contrast to MoEs, which rely on larger backbones, apply disjoint experts in limited layers, and overlook label co-occurrence, \namex{} exploits temporal locality and co-occurrence patterns to reuse LoRA adapters across frames. Its modular design eliminates the need for routing and joint training, and enables flexible expert composition aligned with structural properties of video streams.

\textbf{Adaptive Video Inference and Token Pruning.} Several recent approaches adapt inference-time computation to match input complexity and edge resource constraints. \textit{TOD} and \textit{ROMA}~\cite{lee2021tod,lee2023roma} dynamically select object detectors based on object size, motion, and estimated accuracy without ground-truth. Other strategies adopt model switching~\cite{zhang2020model,salmani2023reconciling} or learn runtime decision policies~\cite{ghosh2023chanakya} to balance latency and accuracy. \textit{Adaptive Model Streaming}~\cite{khani2021real} avoids full model reloads by streaming updates based on content change. At the architectural level, transformer-based methods reduce computation via token pruning. While \textit{Adaptive Token Sampling}~\cite{fayyaz2022adaptive} and \textit{HiRED}~\cite{Arif_Yoon_Nikolopoulos_Vandierendonck_John_Ji_2025} prune tokens using learned or deterministic attention, other works propose learned~\cite{kim2022learned}, adaptive~\cite{yin2022vit}, and hardware-efficient~\cite{dong2023heatvit} token selection. Complementary efforts explore using tiny models or early-exit cascades for efficient routing~\cite{wang2024tiny,park2015big}.

Unlike token pruning or model switching, our approach adapts at the class level by composing LoRA adapters over a shared backbone, avoiding architectural changes or full-model duplication, and achieving higher space efficiency by adding only a small number of parameters as specialized model overhead.

\textbf{Context-aware Classification.} 
CACTUS~\cite{rastikerdar2024cactus} boosts efficiency by switching among small classifiers trained on label subsets, but supports only one active model at a time which limits the performance with overlapping labels. Our method uses composable LoRA adapters to cover multiple labels with lower switching and memory overhead. AdaCon~\cite{neseem2021adacon} builds separate detection heads for co-occurring labels, requiring retraining and added memory. In contrast, our lightweight LoRA adapters share a backbone, improving scalability. Unlike AdaCon’s image-level focus, we leverage temporal continuity for stream-aware adaptation. Related video methods like Uni-AdaFocus~\cite{wang2024uni} and SPMTrack~\cite{cai2025spmtrack} adapt spatial-temporally, but focus on dynamic frame/expert selection rather than label-wise composition.

\begin{figure}[t]
    \centering
    \includegraphics[width=0.6\linewidth]{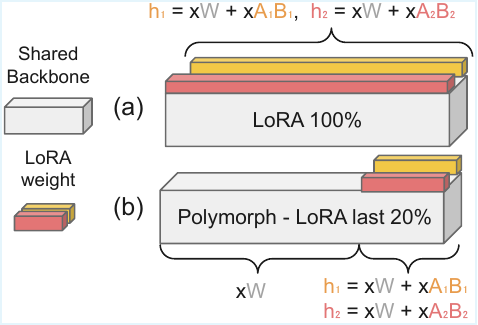}
    \caption{S-LoRA vs Polymorph architecture: (a) applying LoRA to all layers (baseline); (b) Polymorph applies LoRA only to the last 20\% of the layers. Colored bars indicate LoRA adapters.}
    \label{fig:shared-network}
    \vspace{-1em}
\end{figure}

\section{Polymorph}
\label{sec:polymorph}

\textit{Polymorph} is a context-aware system for efficient multi-label video classification. Instead of relying on a single large model, it activates a small set of specialized LoRA adapters tailored to the active labels in each frame. This requires carefully selecting compact label groups, training efficient adapters, and dynamically detecting correct contexts. We address these challenges through: (i) partial LoRA specialization over a shared backbone (Section~\ref{sec:lora-based-context}), (ii) a modular system for adaptive inference (Section~\ref{sec:system-design}), (iii) an optimization formulation for context coverage and reuse (Section~\ref{subsec:problem-formulation}), and (iv) clustering-based training and Inference time context detection algorithms (Sections~\ref{subsec:training-time-context-selection} and~\ref{subsec:runtime-context-detection}).

\subsection{LoRA-based Context Specialization}
\label{sec:lora-based-context}


\begin{figure}[t]
    \centering
    \includegraphics[width=\linewidth]{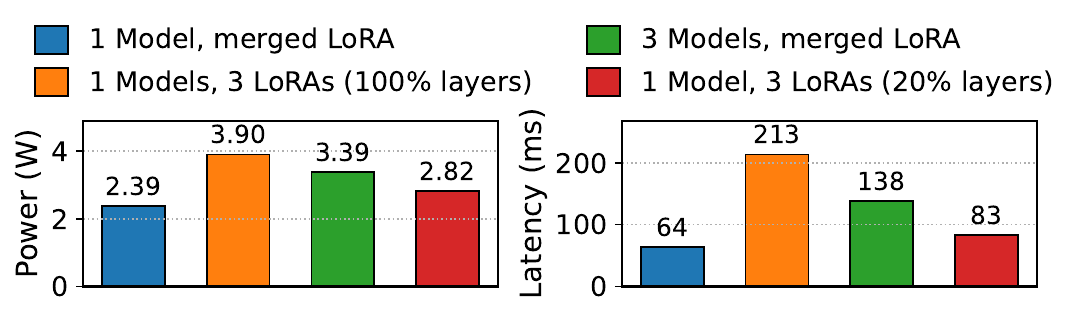}
    \caption{
Comparison of adaptation strategies with three active classifiers. Polymorph applies LoRA adapters only to the final layers of a shared backbone, avoiding other methods overhead.
    }
    \label{fig:lora_energy}
\vspace{-0.5em}
\end{figure}

\begin{figure}[t]
    \centering
    \includegraphics[width=0.9\linewidth]{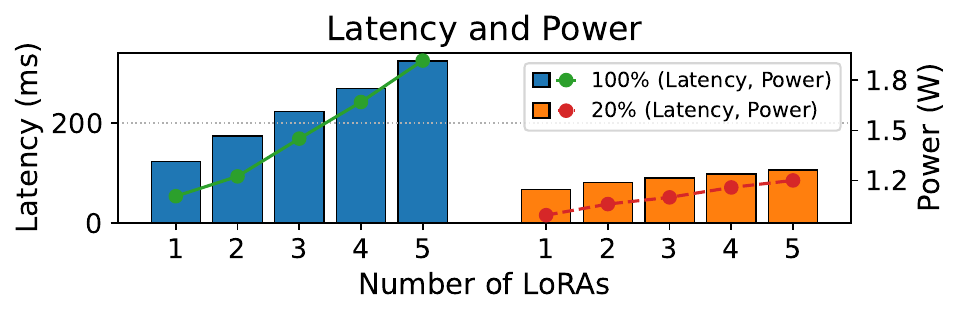}
    \caption{Latency and power as the number of active LoRA adapters increases. Full-model LoRAs (blue and green) incur higher costs per additional adapter. Polymorph (orange and red), which adapts only the final layers, incurs minimal additional cost.}
    \label{fig:lora_count}
    \vspace{-1em}
\end{figure}

\begin{figure*}[t]
    \centering
    \includegraphics[width=0.7\linewidth]{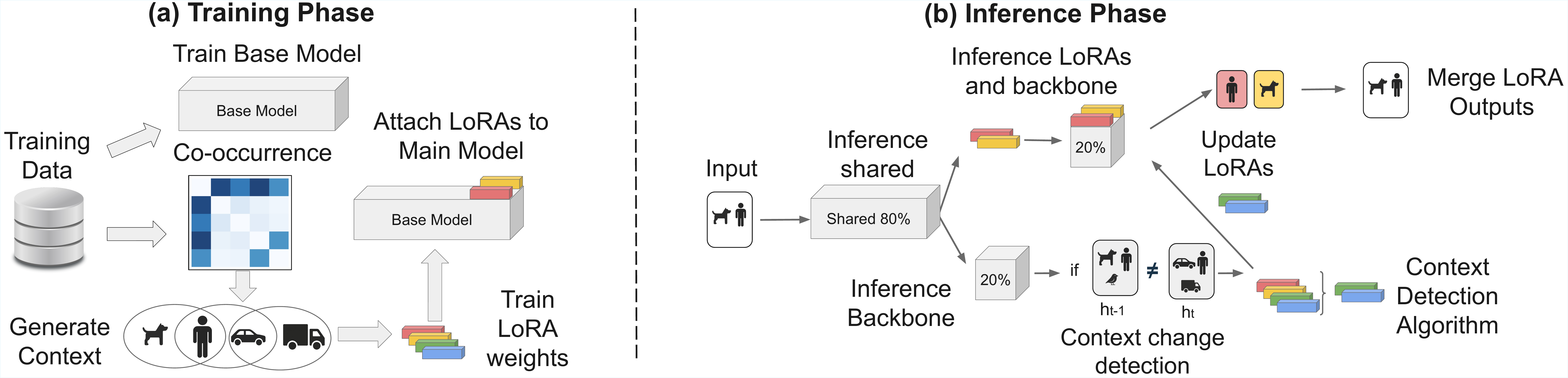}

    \caption{Overview of the \namex{} system architecture. (a) Training: context-specific LoRA adapters are trained based on label co-occurrence. (b) Inference: input passes through a shared backbone and is processed in parallel by the base and context-specific LoRA heads. The base output is compared with the previous frame to detect context changes. If a change is detected, the context detection algorithm updates the selected LoRA adapters. All active LoRA outputs are merged to generate the final results.
}
    \label{fig:system-design}
\vspace{-1em}
\end{figure*}

To support efficient specialization without duplicating entire models, \textit{Polymorph} uses Low-Rank Adaptation (LoRA)~\cite{hu2022lora}, a parameter-efficient fine-tuning method that freezes the weights of a pre-trained base model and injects a pair of small trainable matrices into selected layers. Formally, for a weight matrix \( W \in \mathbb{R}^{h \times d} \), LoRA introduces an update of the form \( W' = W + AB \), where \( A \in \mathbb{R}^{h \times r} \), \( B \in \mathbb{R}^{r \times d} \), and \( r \ll \min(h, d) \). This structure enables efficient specialization using only a small number of additional parameters. In typical usage, LoRA adapters are merged into the backbone weights after training, resulting in a separate model per task or context. While this removes runtime overhead for single-task inference, it limits flexibility in scenarios where multiple concurrent LoRA outputs are needed. To address this, \textit{Polymorph} adopts a strategy inspired by S-LoRA~\cite{sheng2024s-lora}, which avoids merging and instead rewrites the forward pass as \( h = xW' = x(W + AB) = xW + xAB \), allowing the model to dynamically apply multiple LoRA adapters in parallel without altering the base weights. As shown in Figure~\ref{fig:shared-network}a, the backbone computes \( xW \), and selected LoRA outputs \( xA_1B_1, xA_2B_2, \ldots \) are summed on top without merging into backbone. To further improve efficiency, we apply LoRA only to the final 20\% of transformer layers, where class-specific features are most prominent. This additional optimization reduces computational and memory overhead with negligible accuracy loss in our setting (within 1-2 \% per LoRA). Figure~\ref{fig:shared-network}b illustrates this variant: the majority of the model executes once as a shared backbone, and only the final layers are augmented with context-specific LoRA.

Figure~\ref{fig:lora_energy} shows how different competing designs affect power and latency when three context-specific classifiers are active. Details of the experimental setup are provided in Section~\ref{sec:experiments}. Applying three LoRA adapters across all layers of a model without merging weights (orange bars) results in the highest latency (213 ms) and power consumption (3.90 W), as every layer must compute both the backbone and separate LoRA weights. Merging LoRA weights into the model (green bars) reduces power and latency slightly, however this requires instantiation of 3 distinct models. In contrast, Polymorph activates LoRA adapters only in the final layers of a shared backbone (red bars). The early layers are processed once as weights are shared. This design avoids the majority of the overhead of specializing all layers (orange bars) with less than 2\% accuracy degradation.

Figure~\ref{fig:lora_count} further shows how latency and power scale with the number of active LoRA adapters. In full-layer setting, both metrics increase sharply as more adapters are added. In contrast, Polymorph introduces minimal cost even with five active adapters, making it highly efficient for multi-label video inference where multiple contexts may be required.

\subsection{System Design}
\label{sec:system-design}
Polymorph consists of a training phase for constructing context-specific classifiers and an inference phase for adaptive inference (Figure~\ref{fig:system-design}). During the training phase, label co-occurrence patterns are extracted from training data to form compact label groups. Each group is used to train a LoRA adapter. The backbone is trained on all present classes.
At Inference time (Figure~\ref{fig:system-design}(b)), each frame is first passed through the shared backbone, which produces predictions over the full label space. These predictions are used to monitor the current label distribution. If the labels predicted by the shared backbone remain consistent with the previous frame, the system reuses active LoRA adapters. However, if new labels appear, a \textit{context change} is detected. The system then selects a new minimal set of LoRA adapters that together cover the updated label set while satisfying per-class accuracy constraints. These adapters are applied concurrently on top of the shared backbone output, allowing the system to reconfigure efficiently without recomputing shared features or incurring switching overhead.

\subsection{Problem Formulation}
\label{subsec:problem-formulation}

Our goal is to optimize multi-label classification on video streams by activating only a few label-specific LoRA adapters per frame. This reduces computation and switching overhead while maintaining accuracy. Instead of using a single large model, Polymorph applies a compact set of adapters over a shared backbone. The objective function balances two competing goals. The first term minimizes the number of active contexts per frame to reduce memory and energy usage. The second term penalizes changes in the selected context set between consecutive frames to reduce switching overhead and improve temporal stability.

Let $\mathcal{L} = \{1, 2, \dots, K\}$ be the set of all possible labels, and let $\{x_1, x_2, \dots, x_T\}$ be a sequence of video frames. For each frame $x_t$, let $Y_t \subseteq \mathcal{L}$ be the set of active labels in that frame. Our goal is to construct a collection of label subsets that we call contexts: $\mathcal{C} = \{C_1, C_2, \dots, C_M\}$. For each frame $x_t$, we select a small subset of contexts $\mathcal{S}_t \subseteq \mathcal{C}$ such that the union of their labels covers the active labels, $\bigcup_{C \in \mathcal{S}_t} C \supseteq Y_t$. We refer to $\mathcal{S}_t$ as the \textit{active context set} for frame $t$. Let $z_{t,C} \in \{0,1\}$ be an indicator variable denoting whether context $C \in \mathcal{C}$ is in the active context set $\mathcal{S}_t$, i.e., $z_{t,C} = 1$ if $C \in \mathcal{S}_t$, and $z_{t,C} = 0$ otherwise. Our optimizations goals becomes as follows:

\begin{itemize}
    \item The number of selected contexts per frame, $|\mathcal{S}_t|$, is minimized to reduce energy usage (Eq. \ref{eq:ilp_obj} 1st term).
    \item Selected contexts stay similar across adjacent frames to reduce context-switching, $\mathcal{S}_t \approx \mathcal{S}_{t-1}$ (Eq. \ref{eq:ilp_obj} 2nd term).
\end{itemize}

\noindent Subject to the following primary constraint:

\begin{itemize}

    \item Each label $l \in Y_t$ must appear in at least one selected context $C \in \mathcal{S}_t$ where accuracy $a_{C,l}$ is greater than or equal to threshold $\tau$ (based on training data) (Eq. \ref{eq:ilp_cov}).
    \item Context $C_i$ includes at most $B$ labels: $|C_i| \leq B$ (Eq. \ref{eq:ilp_size}).
    \item Total number of contexts is at most $M_{\text{max}}$ (Eq. \ref{eq:ilp_maxC}).
\end{itemize}

\noindent This can be formulated as an Integer Linear Program (ILP):

\vspace{-1em}

{\small
\begin{align}
\text{minimize} \quad & \sum_{t=1}^{T} \left( \sum_{C \in \mathcal{C}} z_{t,C} + \sum_{C \in \mathcal{C}} |z_{t,C} - z_{t-1,C}| \right) \label{eq:ilp_obj} \\
\text{subject to} \quad & \sum_{C: l \in C,\, a_{C,l} \geq \tau} z_{t,C} \geq 1 \quad \forall t, \forall l \in Y_t \label{eq:ilp_cov} \\
& |C| \leq B \quad \forall C \in \mathcal{C} \label{eq:ilp_size} \\
& |\mathcal{C}| \leq M_{\text{max}} \label{eq:ilp_maxC} \\
& z_{t,C} \in \{0,1\} \label{eq:ilp_binary}
\end{align}
}

Solving this ILP exactly is impractical due to the large number of possible context combinations and real-time constraints. Therefore, we use efficient greedy heuristics for both training and inference in sections \ref{subsec:training-time-context-selection} and \ref{subsec:runtime-context-detection}.

\subsection{Training-Time Context Selection}
\label{subsec:training-time-context-selection}

\begin{algorithm}[t]
    \footnotesize
    \caption{Context Construction}
    \label{alg:training}
    \begin{algorithmic}[1]
        \Procedure{BuildContexts}{$\mathbf{C}$, $B$, $M_{\text{max}}$, $allow\_overlap$}
            \State Initialize empty context set $\mathcal{C}$ and $assigned \gets \emptyset$
            \ForAll{valid labels $l \in \mathcal{L}$}
                \If{not $allow\_overlap$ and $l \in assigned$}
                    \State \textbf{continue}
                \EndIf
                \State Build up to size $B$ cluster around $l$ and top co-occurring neighbors
                \If{cluster is unique and $|\mathcal{C}| < M_{\text{max}}$}
                    \State Add cluster to $\mathcal{C}$
                    \If{not $allow\_overlap$}
                        \State $assigned \gets assigned \cup \text{cluster}$
                    \EndIf
                \EndIf
            \EndFor
            \State Identify uncovered labels
            \ForAll{uncovered labels}
                \State Insert into best-fitting context with space remaining
            \EndFor
            \State \Return $\mathcal{C}$
        \EndProcedure
    \end{algorithmic}
\end{algorithm}

Constructing label contexts for training LoRA adapters is a key component of \namex{}’s training pipeline. The objective~ is to group frequently co-occurring labels into compact, semantically meaningful subsets (contexts), while ensuring that all valid labels are covered and no context exceeds a predefined maximum size $B$. These contexts are used to train specialized classifiers and must cover the labels while computationally efficient. We build a co-occurrence matrix from training set and generate clusters centered around frequently occurring labels by greedily expanding them with their strongest co-occurring neighbors. We support two variants: (1) a non-overlapping strategy where each label appears in at most one context, and (2) an overlapping strategy where labels may belong to multiple contexts to better capture semantic structure and improve coverage flexibility. In both variants, unique clusters are retained to minimize redundancy. After initial clustering, any remaining uncovered labels are inserted into compatible existing contexts without violating the context size constraint.
To evaluate cluster quality, we use the following metrics:

\noindent
\textbf{IntraCoherence} measures the average co-occurrence between label pairs $(i, j)$ within each context, encouraging compact and semantically aligned groups ($\text{co}(i, j)$ denotes the normalized frequency with which label pair $(i, j)$).

\begin{equation}
\small
\label{eq:inter-coherence}
\text{IntraCoherence} = \frac{1}{|\mathcal{C}|} \sum_{C \in \mathcal{C}} \frac{1}{|C|(|C|-1)} \sum_{\substack{i,j \in C \\ i \ne j}} \text{co}(i,j)
\end{equation}

\noindent
\textbf{AvgCoverage} Average number of contexts used per frame; Lower number of contexts leads to lower energy cost.
\begin{equation}
\small
\text{AvgCoverage} = \frac{1}{T} \sum_{t=1}^{T} |\mathcal{S}_t|
\end{equation}

\noindent
\textbf{SwitchPenalty} penalizes changes in active contexts between consecutive frames to reduce overhead and context change detection and reclassification errors. 
\begin{equation}
\small
\text{SwitchPenalty} = \sum_{t=2}^{T} \mathbbm{1}[\mathcal{S}_t \ne \mathcal{S}_{t-1}]
\end{equation}

We solve this optimization problem approximately using greedy heuristics. Training constructs compact label groupings based on co-occurrence (Section~\ref{subsec:training-time-context-selection}), while at runtime it selects accurate, stable context subsets per frame using a greedy strategy (Section~\ref{subsec:runtime-context-detection}). This heuristic indirectly optimizes the
\texttt{IntraCoherence} metric explained in Eq. \ref{eq:inter-coherence} and produces a practical context set that balances covering the labels, compactness, and coverage. Compared to exact combinatorial optimization, this method scales well and provides high-quality contexts for downstream deployment.

\subsection{Inference-Time Context Detection}
\label{subsec:runtime-context-detection}

\begin{algorithm}[t]
\footnotesize
\caption{Greedy Context Detection with Conditional Context Copy}
\label{alg:runtime}
\begin{algorithmic}[1]
    \Procedure{DetectContexts}{$Y_t$, $Y_{t-1}$, $\mathcal{C}$, $a_{C,l}$, $\tau$, $\mathcal{S}_{t-1}$, $context\_copy$}
        \If{$Y_t = Y_{t-1}$}
            \State $\mathcal{S}_t \gets \mathcal{S}_{t-1}$
        \ElsIf{$context\_copy$ \textbf{and} $\forall l \in Y_t,\ \exists C \in \mathcal{S}_{t-1}: a_{C,l} \geq \tau$}

            \State $\mathcal{S}_t \gets \mathcal{S}_{t-1}$
        \Else
            \State Initialize $\mathcal{S}_t \gets \emptyset$, $\mathcal{U} \gets Y_t$
            \ForAll{context $C \in \mathcal{C}$ in order of increasing size}
                \State $provides\_labels \gets \{l \in C \cap \mathcal{U} \mid a_{C,l} \geq \tau\}$
                \If{$provides\_labels \neq \emptyset$}
                    \State $\mathcal{S}_t \gets \mathcal{S}_t \cup \{C\}$
                    \State $\mathcal{U} \gets \mathcal{U} \setminus provides\_labels$
                    \If{$\mathcal{U} = \emptyset$}
                        \State \textbf{break}
                    \EndIf
                \EndIf
            \EndFor
        \EndIf
        \State \Return $\mathcal{S}_t$
    \EndProcedure
\end{algorithmic}
\end{algorithm}

At inference time, \namex{}  uses a small set of context-specific classifiers for each frame that collectively cover the active labels with high accuracy. The selected classifiers are normally the same across frames due to the temporal continuity property of video streams. Sometimes, different object types appear and the selection of specialized classifiers needs to be changed to allow accurate classification of these new objects. Alternatively, some types of objects may no longer appear and classifiers may be unloaded to reduce time and energy. The context is determined based of the classification performed by the base model.

In Algorithm~\ref{alg:runtime}, given a frame $x_t$ and its active label set $Y_t \subseteq \mathcal{L}$, the goal is to select a subset of contexts $\mathcal{S}{t} \subseteq \mathcal{C}$ to use on the next frame such that each label $l \in Y_t$ is included in at least one context $C \in \mathcal{S}{t}$. $Y_t$ is based on the classification inferred for frame $x_t$ by the base model, which covers all classes, albeit with lower accuracy than the specialised classes. The set of contexts remains the same when the base model sees no change in context ($Y_t = Y_{t-1}$), or if context reuse is enabled, when the previous context $\mathcal{S}_{t-1}$ still covers $Y_t$ with sufficient accuracy. Due to the manageable number of contexts in our setting, we keep context reuse deactivated, but it is beneficial in datasets involving a large number of possible contexts.

Additionally, we select LoRAs whose expected accuracy $a_{C,l} \geq \tau$, where $\tau$ is a user-defined threshold. These accuracy values \( a_{C,l} \) are computed offline during training by evaluating each context-specific LoRA on a held-out validation set for each of the present classes in the LoRA weight. The threshold \( \tau \) is a parameter that specifies the minimum acceptable accuracy per label and should be set with reference to achievable \( a_{C,l} \) values. Algorithm~\ref{alg:runtime} presents a greedy heuristic that selects a minimum set of contexts that covers all labels in $Y_t$. Priority is given to LoRAs covering fewer classes as these generally have higher accuracy.

The threshold $\tau$ has an impact on the performance of Polymorph. Higher thresholds may increase the size of $\mathcal{S}_t$, which would increase power consumption and latency. Setting $\tau$ too high may render some classes undetectable.

\begin{figure*}[t]
    \centering
    \includegraphics[width=0.9\linewidth]{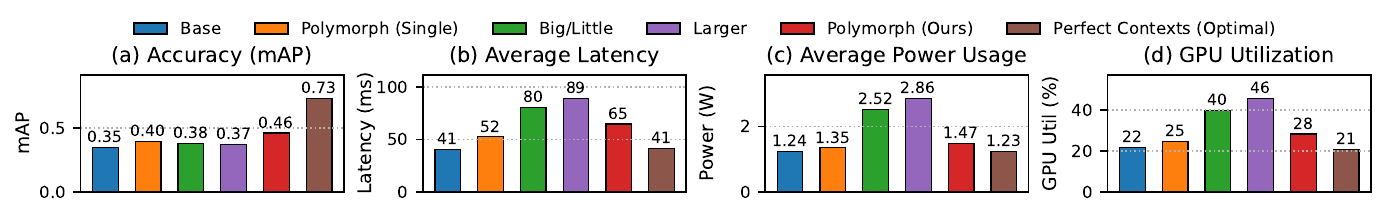}
    \caption{
        Performance comparison across methods (GPU evaluation). 
        \textbf{(a)} Accuracy (mAP) shows that \textit{Polymorph (Ours)} significantly outperforms the \textit{Base}, \textit{Big/Little}, and \textit{Larger} models, approaching the upper bound set by the oracle \textit{Perfect Contexts}. 
        \textbf{(b)} Average latency shows \textit{Polymorph (Ours)} maintains low inference delay, comparable to \textit{Polymorph (Single)}, and faster than \textit{Big/Little} and \textit{Larger}. 
        \textbf{(c)} Average power usage confirms that \textit{Polymorph (Ours)} is more energy-efficient than all other deployable methods, while delivering significantly higher accuracy. Latency, power, and GPU utilization remain similar across variants with shared backbone usage.
    }
    
    \label{fig:comparison}
\vspace{-0.5em}
\end{figure*}

\section{Experimental Evaluation}
\label{sec:experiments}

We implemented \namex{} in 3k python code. The LoRA adapters of \namex{} were implemented on top of the Hugging Face PEFT library \cite{peft}. We modified the library to implement shared layers for the initial layer of the neural network as explained in Section \ref{sec:lora-based-context}. The target embedded device that we used was a Jetson Orin Nano, which features a 6-core ARM Cortex-A78AE CPU, a 1024-core Ampere GPU with 32 Tensor Cores, and 8 GB of LPDDR5 memory. We used the Jetson Stats tool \cite{bonghi_jetson_stats_2025} to measure energy consumption and monitor performance during runtime.

We compare \namex{} with the following baselines. \textbf{\namex{} (Single)}, A constrained version of \namex{} activates only one context at a time, disabling concurrent adapters and selecting a single context per frame. This setup behaves similarly to \textbf{CACTUS}~\cite{rastikerdar2024cactus}, which switches between separate micro-classifiers using hard context boundaries. We also compare two ViT models of different sizes: DeiT-tiny, used as the base model in our method and trained on all classes, and a larger ViT trained on the same full label set. Additionally, we include a \textbf{Big/Little} baseline, similar to prior work~\cite{park2015big, lee2021tod, wang2024tiny}, where a small model is used by and a larger model is queried only when predictions may be uncertain. We use a context change detector to identify such cases and trigger the larger model when needed. As an upper bound, we report performance of \textbf{Perfect Contexts} baseline that always selects the optimal set of LoRA adapters per frame using ground-truth label information.

\begin{table}[t]
    \centering
    \small
    \caption{Validation performance on COCO using subset accuracy, F1 (micro/macro), and mAP. LoRA scores are averaged.}
    \label{tab:accuracy-results}
    \resizebox{\columnwidth}{!}{%
    \begin{tabular}{lcccc}
        \toprule
        \textbf{Model} & \textbf{Subset Acc. (\%)} & \textbf{F1 (Micro)} & \textbf{F1 (Macro)} & \textbf{mAP} \\
        \midrule
        DeiT-tiny (base) & 31.3 & 67.6 & 62.4 & 44.0 \\
        ViT-large        & 41.3 & 76.7 & 73.3 & 57.7 \\
        LoRA (averaged)  & 65.3 & 81.6 & 76.6 & 66.0 \\
        \bottomrule
    \end{tabular}
    }
    \vspace{-1em}
\end{table}

We trained on the MS COCO dataset~\cite{lin2014microsoft}, which provides over 330{,}000 images with multi-label annotations across 80 categories. For video-time inference, we used the TAO dataset~\cite{dave2020tao}, which contains over 2{,}900 high-resolution video segments with frame-level annotations for more than 800 object classes. Due to its size, complex annotations, and high computational cost, we did not use TAO for training. Its sparse and imbalanced label distribution makes it a valuable benchmark for evaluating context-aware methods like \namex{}. To simulate a realistic 5~fps video feed, we introduced timed delays in the inference loop to enforce intervals between processed frames. Evaluating the full TAO validation set on embedded devices is prohibitively time-consuming (taking several days per run). Therefore, we selected a representative subset. Using the smallest base model, we computed per-video mAP and retained only videos with accuracy above 0.2 containing 119 labels per frame, ensuring selected clips contained sufficient label signal to meaningfully evaluate context-awareness.

The base model was trained on the MS COCO dataset using the DeiT-tiny architecture~\cite{touvron2021training} for 60 epochs. The model was optimized with SGD using a learning rate of 0.001 and a batch size of 8. All models used pre-trained weights and were trained on a RTX 6000 GPU on Chameleon Cloud \cite{keahey2020lessons}. For the LoRA weights, we trained separate models for 3 epochs using the same optimizer settings but with a reduced batch size of 8. Each LoRA model was trained only on images that contained the specific classes it was responsible for detecting. The LoRA adapters were injected into the \texttt{query} and \texttt{value} projections of the attention blocks, with a rank of 16, scaling factor of 32, and dropout of 0.1. 
For each LoRA, a classification head was fine-tuned during training alongside the LoRA weights, which are only present at the last 20\% of the layers. Table~\ref{tab:accuracy-results} reports multi-label classification training metrics and better LoRA. Subset accuracy reflects exact match per instance, F1 scores summarize precision-recall trade-offs, and mAP captures label ranking quality. LoRA weights achieve strong improvements.

\begin{table}[t]
\centering
\footnotesize
\begin{tabular}{lrrr}
\toprule
Method & Params (M) & Memory (Mb) & MACs (M) \\
\midrule
Larger & 85.86 & 327.54 & 16866.65 \\
Base & 5.54 & 21.14 & 1079.39 \\
Polymorph & 5.82 & 22.21 & 1084.23 \\
\bottomrule
\end{tabular}
\caption{Comparison of model size metrics for each method.}
\label{tab:model_stats}
\vspace{-1em}
\end{table}

\begin{figure}[t]
    \centering    \includegraphics[width=0.35\textwidth]{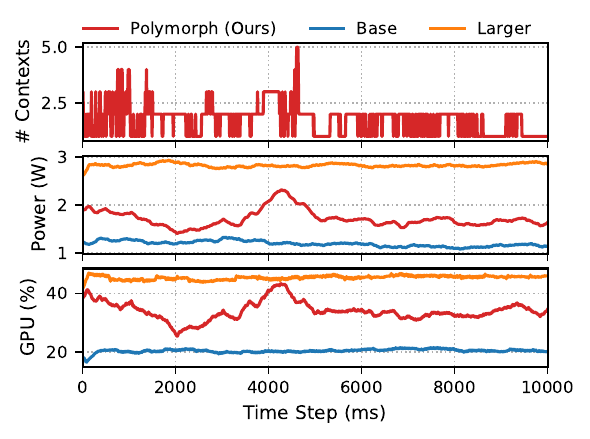}
    \caption{
    Temporal profile of context usage (top), power consumption (middle), and GPU utilization (bottom) for \textit{Polymorph (Ours)} compared to \textit{Base} and \textit{Larger} models. 
    }
    \label{fig:context-temporal}

\vspace{-1.5em}
\end{figure}

Figure~\ref{fig:comparison} presents a GPU-based comparison across baselines in terms of accuracy, latency, and energy consumption. \textit{Polymorph (Ours)} achieves a mean Average Precision (mAP) of 0.46 (Figure~\ref{fig:comparison}a), outperforming both the \textit{Base} (0.35), \textit{Big/Little} (0.38), and \textit{Larger} (0.37) models, despite using significantly fewer parameters and lower compute, as shown in Table~\ref{tab:model_stats}. This demonstrates that context-aware specialization using LoRA adapters can offer higher discriminative performance than monolithic classification or heuristic switching strategies. Compared to \textit{Big/Little}, which selectively triggers a larger model under uncertainty, \textit{Polymorph} delivers higher accuracy without incurring the cost of model switching. The \textit{Perfect Contexts} baseline, which assumes access to an oracle context selector at each frame, achieves an upper-bound mAP of 0.73, suggesting significant remaining potential with improved context detection. In terms of latency (Figure~\ref{fig:comparison}b), \textit{Polymorph} sustains real-time throughput at 41\,ms per frame, closely matching the performance of \textit{Polymorph (Single)}, and remaining significantly faster than \textit{Larger} (89\,ms) and \textit{Big/Little} (80\,ms), both of which rely on heavier models. This confirms that concurrent execution of LoRA adapters introduces minimal delay when operating on a shared backbone. Power measurements (Figure~\ref{fig:comparison}c) further emphasize \textit{Polymorph}'s efficiency: it consumes only 1.47\,W on average, compared to 2.52\,W for \textit{Big/Little} and 2.86\,W for \textit{Larger}. Latency, power consumption, and GPU utilization remain similar across Polymorph variants due to the reuse of early backbone layers and the lightweight nature of LoRA adapters. Overall, \textit{Polymorph} achieves favorable trade-offs across all evaluation metrics in the GPU setting, offering significantly better accuracy than existing deployable methods, while using less energy and maintaining low latency. Figure~\ref{fig:context-temporal} further supports this by showing that increases in the number of active contexts in \textit{Polymorph} lead to higher power draw and GPU utilization. To limit this overhead, the context detection algorithm (Algorithm~\ref{alg:runtime}) selects the smallest set of contexts that meet the required accuracy threshold. This is why \textit{Polymorph} prioritizes using fewer contexts whenever possible, and keeping energy and GPU usage low.

\begin{table}[t]
    \centering
    \footnotesize
    \caption{Effect of maximum context size $B$ and count $|\mathcal{C}|$ on mAP. Comparing two variants of Algorithm~\ref{alg:runtime} that reuse LoRAs from the last selected context (context copy) vs \namex{}.}
    \label{tab:context_map}
    \begin{tabular}{cccc}
        \toprule
        \multicolumn{1}{c}{\small $|\mathcal{C}|$ × $B$} & \textbf{Oracle} & \textbf{Context Copy} & \textbf{\namex{}} \\
        \midrule
        26×2  & 78.7 & 44.5 & 48.0 \\
        18×3  & 75.3 & 43.8 & 46.6 \\
        11×5  & 73.5 & 44.2 & 46.2 \\
        6×10  & 59.7 & 43.6 & 45.6 \\
        4×15  & 54.7 & 44.3 & 44.8 \\
        3×20  & 52.8 & 43.9 & 44.4 \\
        \bottomrule
    \end{tabular}
    \vspace{-1em}
\end{table}

Table~\ref{tab:context_map} evaluates the effect of maximum context size $B$ and number of generated contexts using that, $|\mathcal{C}|$ on classification accuracy (mAP) under different context selection strategies. E.g., 11×5 indicates 11 contexts each containing up to five labels. We compare an oracle that selects the best context set per frame (Perfect contexts in Figure \ref{fig:comparison}) with two Polymorph variants: one that copies the previous frame’s context set if they can cover the labels (\texttt{context\_copy} set to true in Algorithm \ref{alg:runtime}), and one that dynamically re-selects contexts using Algorithm~\ref{alg:runtime} which is the one we used as main \namex{}. \namex{} consistently outperforms copying across all configurations, since at every iteration of the algorithm it tries to find the best context set rather than retaining the existing ones. The gap is small due to a relatively small number of significant changes of video content. Among the tested settings, 11×5 achieves the best trade-off between compactness and accuracy, reaching 46.2 mAP while maintaining moderate context granularity. The gap between the Oracle results and the \namex{} results indicates that small contexts (few labels per context) can achieve high accuracy, in part due to higher per-LoRA accuracy (Figure~\ref{fig:larger-smaler-models}). However, detecting the context is harder when contexts have few labels (small $B$). A first challenge is detecting what labels might be occurring in the video frames in the absence of ground truth data. This problem is easier with large $B$ as such classifiers can detect many object types even when the base model observed few. Our preliminary investigation using a larger model as the all-class classifier did not yield a significant improvement in end-to-end mAP, leaving the discovery of potential labels as an open issue. A second challenge relates to using better heuristics rather than greedy strategies.

Table~\ref{tab:clustering_metrics} analyzes the clustering quality of different context construction algorithms used during training (Algorithm~\ref{alg:training}). \namex{} is the version that does not allow overlapping classes across contexts. The basic method randomly assigns labels to same-sized clusters, while overlap is the \namex{} variant allowing overlapping labels across clusters. We report three metrics introduced in Section~\ref{subsec:problem-formulation}: \texttt{IntraCoherence}, which measures semantic alignment within contexts; \texttt{AvgCoverage}, the number of contexts needed per frame; and \texttt{SwitchPenalty}, the frequency of context changes. As context size decreases, the overlap-based method significantly improves coherence (e.g., 2258.8 at size 5) while also reducing coverage and switching overhead (1.07 contexts/frame and 1417 switches). Based on this analysis, we selected the size-5 setting for our experiments, as it offers the best trade-off between these metrics and a manageable number of context-specific LoRA weights. The overlapping method has the best theoretical performance for the metrics; however, due to the large number of contexts it requires (e.g., 45 vs. 11 at size 5) and its higher susceptibility to context detection errors, we used the non-overlapping method as \namex{}.

\begin{table}[t]
    \centering
    \footnotesize
    \caption{Clustering metrics across algorithms and configurations. ↓ = lower is better. context tuple: \#contexts × max context size.}
    \label{tab:clustering_metrics}
    \begin{tabular}{llccc}
        \toprule
        \textbf{$|\mathcal{C}|$ × $B$} & \textbf{Algorithm} & \textbf{Intra ↓} & \textbf{Coverage ↓} & \textbf{Switches ↓} \\
        \midrule
        26×2  & \textbf{basic}       & 524.7  & 1.3923 & 2004 \\
        26×2  & \textbf{\namex{}}   & 915.6  & 1.3614 & 1664 \\
        50×2  & \textbf{overlap}     & 2318.4 & 1.1674 & 1594 \\
        \cline{1-5}
        11×5  & \textbf{basic}       & 472.0  & 1.3355 & 1594 \\
        11×5  & \textbf{\namex{}}   & 693.9  & 1.3300 & 1522 \\
        45×5  & \textbf{overlap}     & 2258.8 & 1.0735 & 1417 \\
        \cline{1-5}
        4×15  & \textbf{basic}       & 348.5  & 1.2548 & 1151 \\
        4×15  & \textbf{\namex{}}   & 424.0  & 1.1551 & 758  \\
        50×15 & \textbf{overlap}     & 925.2  & 1.0057 & 688  \\
        \bottomrule
    \end{tabular}
\end{table}

\begin{figure}[t]
    \centering
    \includegraphics[width=0.5\textwidth]{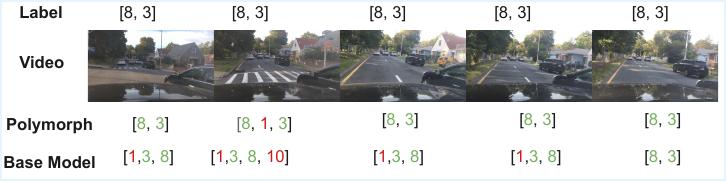}
    \caption{Polymorph avoids false positives by using narrower classifiers, unlike the base model which predicts non-existent classes (e.g., 1, 10). 1: Person, 3: Car, 8: Truck, 10: Traffic Light}
    \label{fig:accuracy-results}
    \vspace{-1em}
\end{figure}

Figure~\ref{fig:accuracy-results} illustrates the effect of false negatives in a single video sequence. The ground truth labels remain constant across frames, yet the base model frequently predicts non-existent classes (e.g., 1, 10), leading to false positives. In contrast, Polymorph maintains accurate predictions throughout. Due to using narrower, context-specific classifiers that restrict the label space per frame, reducing the chance of spurious detections. By focusing on relevant label subsets, Polymorph avoids activating unrelated classes.

\section{Conclusion and Future Work}

We introduced \textbf{\namex{}}, a context-aware multi-label video classification method for efficient on-device inference. By leveraging label sparsity, temporal continuity, and co-occurrence, Polymorph partitions the label space into context-specific LoRA adapters. These are dynamically composed at runtime to match active labels, enabling accurate, low-latency inference under resource constraints. On the TAO benchmark, Polymorph reduces energy by 40\%, and improves mAP by 9 points compared to baselines. 
Future work includes runtime adaptation to distribution shifts by detecting context changes and dynamically updating context assignments when co-occurrence patterns evolve; automatic detection of scene shifts or label drift using online monitoring; and hybrid cloud-edge setups in which LoRA adapters are periodically retrained or re-clustered in the cloud based on streaming video data, then pushed to edge devices.

\section*{Acknowledgements}

This work was supported by a research grant from the Department for the Economy, Northern Ireland, under the US-Ireland R\&D Partnership Programme under grant agreement USI-226, by Science Foundation Ireland, grant agreement 22/US/3848, and by the National Science Foundation under Grant No.\ 2315851. Results presented in this paper were obtained using the Chameleon testbed supported by the National Science Foundation.

{\small
\bibliographystyle{ieee_fullname}
\bibliography{egbib}

@String(MobiSys = {Proc. Int. Conf. Mobile Syst., Appl., and Services})

@String(ICFEC = {IEEE Int. Conf. Fog Edge Comput.})

@String(ICCV = {Int. Conf. Comput. Vis.})

@inproceedings{rastikerdar2024cactus,
  author    = {Mohammad Mehdi Rastikerdar and Jin Huang and Shiwei Fang and Hui Guan and Deepak Ganesan},
  title     = {{CACTUS}: Dynamically Switchable Context-aware micro-Classifiers for Efficient {IoT} Inference},
  booktitle = MobiSys,
  pages     = {505--518},
  year      = {2024}
}

@inproceedings{lee2021tod,
  author    = {JunKyu Lee and Blesson Varghese and Roger Woods and Hans Vandierendonck},
  title     = {{TOD}: Transprecise Object Detection to Maximise Real-Time Accuracy on the Edge},
  booktitle = ICFEC,
  pages     = {53--60},
  year      = {2021},
  organization = {IEEE}
}

@inproceedings{khani2021real,
  author    = {Mehrdad Khani and Pouya Hamadanian and Arash Nasr-Esfahany and Mohammad Alizadeh},
  title     = {Real-Time Video Inference on Edge Devices via Adaptive Model Streaming},
  booktitle = ICCV,
  pages     = {4572--4582},
  year      = {2021}
}

@article{ke2020surveillance,
  author  = {Ruimin Ke and Yifan Zhuang and Ziyuan Pu and Yinhai Wang},
  title   = {A Smart, Efficient, and Reliable Parking Surveillance System with Edge Artificial Intelligence on IoT Devices},
  journal = {IEEE Transactions on Intelligent Transportation Systems},
  volume  = {22},
  number  = {8},
  pages   = {4962--4974},
  year    = {2020}
}

@article{groshev2023robotics,
  title={Edge robotics: Are we ready? An experimental evaluation of current vision and future directions},
  author={Groshev, Milan and Baldoni, Gabriele and Cominardi, Luca and de la Oliva, Antonio and Gazda, Robert},
  journal={Digital Communications and Networks},
  volume={9},
  number={1},
  pages={166--174},
  year={2023},
  publisher={Elsevier}
}

@article{incel2023health,
  author  = {Ozlem Durmaz Incel and Sevda {\"O}zge Bursa},
  title   = {On-Device Deep Learning for Mobile and Wearable Sensing Applications: A Review},
  journal = {IEEE Sensors Journal},
  volume  = {23},
  number  = {6},
  pages   = {5501--5512},
  year    = {2023}
}

@article{cheng2024pruning,
  author  = {Hongrong Cheng and Miao Zhang and Javen Qinfeng Shi},
  title   = {A Survey on Deep Neural Network Pruning: Taxonomy, Comparison, Analysis, and Recommendations},
  journal = {IEEE Transactions on Pattern Analysis and Machine Intelligence},
  year    = {2024}
}

@article{rokh2023quantization,
  author  = {Babak Rokh and Ali Azarpeyvand and Alireza Khanteymoori},
  title   = {A Comprehensive Survey on Model Quantization for Deep Neural Networks in Image Classification},
  journal = {ACM Transactions on Intelligent Systems and Technology},
  volume  = {14},
  number  = {6},
  pages   = {1--50},
  year    = {2023}
}

@article{ren2021nas,
  author  = {Pengzhen Ren and Yun Xiao and Xiaojun Chang and Po-Yao Huang and Zhihui Li and Xiaojiang Chen and Xin Wang},
  title   = {A Comprehensive Survey of Neural Architecture Search: Challenges and Solutions},
  journal = {ACM Computing Surveys (CSUR)},
  volume  = {54},
  number  = {4},
  pages   = {1--34},
  year    = {2021}
}

@article{hu2022lora,
  author  = {Edward J. Hu and Yelong Shen and Phillip Wallis and Zeyuan Allen-Zhu and Yuanzhi Li and Shean Wang and Lu Wang and Weizhu Chen},
  title   = {LoRA: Low-Rank Adaptation of Large Language Models},
  journal = {International Conference on Learning Representations (ICLR)},
  volume  = {1},
  number  = {2},
  pages   = {3},
  year    = {2022}
}

@inproceedings{neseem2021adacon,
  author    = {Marina Neseem and Sherief Reda},
  title     = {AdaCon: Adaptive Context-Aware Object Detection for Resource-Constrained Embedded Devices},
  booktitle = {IEEE/ACM International Conference on Computer Aided Design (ICCAD)},
  pages     = {1--9},
  year      = {2021}
}

@inproceedings{aghapour2024piqi,
  title={Piqi: Partially quantized DNN inference on HMPSoCs},
  author={Aghapour, Ehsan and Shen, Yixian and Sapra, Dolly and Pimentel, Andy and Pathania, Anuj},
  booktitle={Proceedings of the 29th ACM/IEEE International Symposium on Low Power Electronics and Design},
  pages={1--6},
  year={2024}
}

@article{zhao2024edge,
  title={Edge-MPQ: Layer-Wise Mixed-Precision Quantization with Tightly Integrated Versatile Inference Units for Edge Computing},
  author={Zhao, Xiaotian and Xu, Ruge and Gao, Yimin and Verma, Vaibhav and Stan, Mircea R and Guo, Xinfei},
  journal={IEEE Transactions on Computers},
  year={2024},
  publisher={IEEE}
}

@inproceedings{shen2024agile,
  title={Agile-quant: Activation-guided quantization for faster inference of LLMs on the edge},
  author={Shen, Xuan and Dong, Peiyan and Lu, Lei and Kong, Zhenglun and Li, Zhengang and Lin, Ming and Wu, Chao and Wang, Yanzhi},
  booktitle={Proceedings of the AAAI Conference on Artificial Intelligence},
  volume={38},
  number={17},
  pages={18944--18951},
  year={2024}
}

@inproceedings{lepikhingshard,
  title={GShard: Scaling Giant Models with Conditional Computation and Automatic Sharding},
  author={Lepikhin, Dmitry and Lee, HyoukJoong and Xu, Yuanzhong and Chen, Dehao and Firat, Orhan and Huang, Yanping and Krikun, Maxim and Shazeer, Noam and Chen, Zhifeng},
  booktitle={International Conference on Learning Representations}
}

@article{fedus2022switch,
  title={Switch transformers: Scaling to trillion parameter models with simple and efficient sparsity},
  author={Fedus, William and Zoph, Barret and Shazeer, Noam},
  journal={Journal of Machine Learning Research},
  volume={23},
  number={120},
  pages={1--39},
  year={2022}
}

@inproceedings{yu20248,
  title={8-bit Transformer Inference and Fine-tuning for Edge Accelerators},
  author={Yu, Jeffrey and Prabhu, Kartik and Urman, Yonatan and Radway, Robert M and Han, Eric and Raina, Priyanka},
  booktitle={Proceedings of the 29th ACM International Conference on Architectural Support for Programming Languages and Operating Systems, Volume 3},
  pages={5--21},
  year={2024}
}

@article{daxberger2023mobile,
  title={Mobile V-MoEs: Scaling Down Vision Transformers via Sparse Mixture-of-Experts},
  author={Daxberger, Erik A and Weers, Floris and Zhang, Bowen and Gunter, Tom and Pang, Ruoming and Eichner, Marcin and Emmersberger, Michael and Yang, Yinfei and Toshev, Alexander and Du, Xianzhi},
  journal={CoRR},
  year={2023}
}

@inproceedings{sarkar2023edge,
  title={Edge-MoE: Memory-efficient multi-task vision transformer architecture with task-level sparsity via mixture-of-experts},
  author={Sarkar, Rishov and Liang, Hanxue and Fan, Zhiwen and Wang, Zhangyang and Hao, Cong},
  booktitle={2023 IEEE/ACM International Conference on Computer Aided Design (ICCAD)},
  pages={01--09},
  year={2023},
  organization={IEEE}
}

@article{Arif_Yoon_Nikolopoulos_Vandierendonck_John_Ji_2025, title={HiRED: Attention-Guided Token Dropping for Efficient Inference of High-Resolution Vision-Language Models}, volume={39}, url={https://ojs.aaai.org/index.php/AAAI/article/view/32171}, DOI={10.1609/aaai.v39i2.32171}, abstractNote={High-resolution Vision-Language Models (VLMs) are widely used in multimodal tasks to enhance accuracy by preserving detailed image information. However, these models often generate an excessive number of visual tokens due to the need to encode multiple partitions of a high-resolution image input. Processing such a large number of visual tokens poses significant computational challenges, particularly for resource-constrained commodity GPUs. To address this challenge, we propose High-Resolution Early Dropping (HiRED), a plug-and-play token-dropping method designed to operate within a fixed token budget. HiRED leverages the attention of CLS token in the vision transformer (ViT) to assess the visual content of the image partitions and allocate an optimal token budget for each partition accordingly. The most informative visual tokens from each partition within the allocated budget are then selected and passed to the subsequent Large Language Model (LLM). We showed that HiRED achieves superior accuracy and performance, compared to existing token-dropping methods. Empirically, HiRED-20% (i.e., a 20% token budget) on LLaVA-Next-7B achieves a 4.7x increase in token generation throughput, reduces response latency by 78%, and saves 14% of GPU memory for single inference on an NVIDIA TESLA P40 (24 GB). For larger batch sizes (e.g., 4), HiRED-20% prevents out-of-memory errors by cutting memory usage by 30%, while preserving throughput and latency benefits.}, number={2}, journal={Proceedings of the AAAI Conference on Artificial Intelligence}, author={Arif, Kazi Hasan Ibn and Yoon, JinYi and Nikolopoulos, Dimitrios S. and Vandierendonck, Hans and John, Deepu and Ji, Bo}, year={2025}, month={Apr.}, pages={1773-1781} }

@inproceedings{fayyaz2022adaptive,
  title={Adaptive token sampling for efficient vision transformers},
  author={Fayyaz, Mohsen and Koohpayegani, Soroush Abbasi and Jafari, Farnoush Rezaei and Sengupta, Sunando and Joze, Hamid Reza Vaezi and Sommerlade, Eric and Pirsiavash, Hamed and Gall, J{\"u}rgen},
  booktitle={European Conference on Computer Vision},
  pages={396--414},
  year={2022},
  organization={Springer}
}

@inproceedings{lee2023roma,
  title={Roma: Run-time object detection to maximize real-time accuracy},
  author={Lee, JunKyu and Varghese, Blesson and Vandierendonck, Hans},
  booktitle={Proceedings of the IEEE/CVF Winter Conference on Applications of Computer Vision},
  pages={6405--6414},
  year={2023}
}

@inproceedings{dosovitskiy2021an,
  title     = {An Image is Worth 16x16 Words: Transformers for Image Recognition at Scale},
  author    = {Alexey Dosovitskiy and Lucas Beyer and Alexander Kolesnikov and Dirk Weissenborn and Xiaohua Zhai and Thomas Unterthiner and Mostafa Dehghani and Matthias Minderer and Georg Heigold and Sylvain Gelly and Jakob Uszkoreit and Neil Houlsby},
  booktitle = {International Conference on Learning Representations (ICLR)},
  year      = {2021},
  url       = {https://openreview.net/forum?id=YicbFdNTTy}
}

@inproceedings{touvron2021training,
  title     = {Training data-efficient image transformers \& distillation through attention},
  author    = {Hugo Touvron and Matthieu Cord and Matthijs Douze and Francisco Massa and Alexandre Sablayrolles and Herv{\'e} J{\'e}gou},
  booktitle = {Proceedings of the 38th International Conference on Machine Learning (ICML)},
  year      = {2021},
  pages     = {10347--10357},
  publisher = {PMLR},
  url       = {http://proceedings.mlr.press/v139/touvron21a.html}
}

@misc{nvidia_tensorrt_2025,
  title        = {{NVIDIA TensorRT: High-Performance Deep Learning Inference SDK}},
  author       = {{NVIDIA Corporation}},
  year         = {2025},
  howpublished = {\url{https://developer.nvidia.com/tensorrt}},
  note         = {Accessed June 2025},
  organization = {NVIDIA Corporation}
}

@misc{onnx_2025,
  title        = {{ONNX: Open Neural Network Exchange}},
  author       = {{ONNX Community}},
  year         = {2025},
  howpublished = {\url{https://onnx.ai/}},
  note         = {Accessed June 2025}
}

@inproceedings{sheng2024s-lora,
  title     = {S‑LoRA: Serving Thousands of Concurrent LoRA Adapters},
  author    = {Ying Sheng and Shiyi Cao and Dacheng Li and Coleman Hooper and Nicholas Lee and Shuo Yang and Christopher Chou and Banghua Zhu and Lianmin Zheng and Kurt Keutzer and Joseph E. Gonzalez and Ion Stoica},
  booktitle = {Proceedings of the 5th Symposium on Systems for Machine Learning (MLSys)},
  year      = {2024},
  address   = {Santa Clara, CA, USA},
  note      = {arXiv:2311.03285},
}

@misc{bonghi_jetson_stats_2025,
  author       = {Raffaello Bonghi},
  title        = {{jetson‑stats and jtop: Monitoring and Control for NVIDIA Jetson Devices}},
  howpublished = {\url{https://developer.nvidia.com/embedded/community/jetson-projects/jetson_stats}},
  year         = {2025},
  note         = {Accessed June 2025; latest PyPI release 4.3.2 (Mar 19, 2025)},
  organization = {NVIDIA Developer / PyPI},
}

@misc{peft,
  title        = {PEFT: State-of-the-art Parameter-Efficient Fine-Tuning methods},
  author       = {Sourab Mangrulkar and Sylvain Gugger and Lysandre Début and Younes Belkada and Sayak Paul and Benjamin Bossan},
  howpublished = {\url{https://github.com/huggingface/peft}},
  year         = {2022}
}

@inproceedings{kim2022learned,
  title={Learned token pruning for transformers},
  author={Kim, Sehoon and Shen, Sheng and Thorsley, David and Gholami, Amir and Kwon, Woosuk and Hassoun, Joseph and Keutzer, Kurt},
  booktitle={Proceedings of the 28th ACM SIGKDD Conference on Knowledge Discovery and Data Mining},
  pages={784--794},
  year={2022}
}

@inproceedings{yin2022vit,
  title={A-vit: Adaptive tokens for efficient vision transformer},
  author={Yin, Hongxu and Vahdat, Arash and Alvarez, Jose M and Mallya, Arun and Kautz, Jan and Molchanov, Pavlo},
  booktitle={Proceedings of the IEEE/CVF conference on computer vision and pattern recognition},
  pages={10809--10818},
  year={2022}
}

@inproceedings{dong2023heatvit,
  title={Heatvit: Hardware-efficient adaptive token pruning for vision transformers},
  author={Dong, Peiyan and Sun, Mengshu and Lu, Alec and Xie, Yanyue and Liu, Kenneth and Kong, Zhenglun and Meng, Xin and Li, Zhengang and Lin, Xue and Fang, Zhenman and others},
  booktitle={2023 IEEE International Symposium on High-Performance Computer Architecture (HPCA)},
  pages={442--455},
  year={2023},
  organization={IEEE}
}

@article{ghosh2023chanakya,
  title={Chanakya: Learning runtime decisions for adaptive real-time perception},
  author={Ghosh, Anurag and Balloli, Vaibhav and Nambi, Akshay and Singh, Aditya and Ganu, Tanuja},
  journal={Advances in Neural Information Processing Systems},
  volume={36},
  pages={55668--55680},
  year={2023}
}

@inproceedings{wang2024tiny,
  title={Tiny Models are the Computational Saver for Large Models},
  author={Wang, Qingyuan and Cardiff, Barry and Frapp{\'e}, Antoine and Larras, Benoit and John, Deepu},
  booktitle={European Conference on Computer Vision},
  pages={163--182},
  year={2024},
  organization={Springer}
}

@inproceedings{salmani2023reconciling,
  title={Reconciling high accuracy, cost-efficiency, and low latency of inference serving systems},
  author={Salmani, Mehran and Ghafouri, Saeid and Sanaee, Alireza and Razavi, Kamran and M{\"u}hlh{\"a}user, Max and Doyle, Joseph and Jamshidi, Pooyan and Sharifi, Mohsen},
  booktitle={Proceedings of the 3rd Workshop on Machine Learning and Systems},
  pages={78--86},
  year={2023}
}

@inproceedings{zhang2020model,
  title={$\{$Model-Switching$\}$: Dealing with fluctuating workloads in $\{$Machine-Learning-as-a-Service$\}$ systems},
  author={Zhang, Jeff and Elnikety, Sameh and Zarar, Shuayb and Gupta, Atul and Garg, Siddharth},
  booktitle={12th USENIX Workshop on Hot Topics in Cloud Computing (HotCloud 20)},
  year={2020}
}

@inproceedings{park2015big,
  title={Big/little deep neural network for ultra low power inference},
  author={Park, Eunhyeok and Kim, Dongyoung and Kim, Soobeom and Kim, Yong-Deok and Kim, Gunhee and Yoon, Sungroh and Yoo, Sungjoo},
  booktitle={2015 international conference on hardware/software codesign and system synthesis (codes+ isss)},
  pages={124--132},
  year={2015},
  organization={IEEE}
}

@inproceedings{cai2025spmtrack,
  title={SPMTrack: Spatio-Temporal Parameter-Efficient Fine-Tuning with Mixture of Experts for Scalable Visual Tracking},
  author={Cai, Wenrui and Liu, Qingjie and Wang, Yunhong},
  booktitle={Proceedings of the Computer Vision and Pattern Recognition Conference},
  pages={16871--16881},
  year={2025}
}

@article{wang2024uni,
  title={Uni-AdaFocus: Spatial-Temporal Dynamic Computation for Video Recognition},
  author={Wang, Yulin and Zhang, Haoji and Yue, Yang and Song, Shiji and Deng, Chao and Feng, Junlan and Huang, Gao},
  journal={IEEE Transactions on Pattern Analysis and Machine Intelligence},
  year={2024},
  publisher={IEEE}
}

@inproceedings{lin2014microsoft,
  title={Microsoft coco: Common objects in context},
  author={Lin, Tsung-Yi and Maire, Michael and Belongie, Serge and Hays, James and Perona, Pietro and Ramanan, Deva and Doll{\'a}r, Piotr and Zitnick, C Lawrence},
  booktitle={Computer vision--ECCV 2014: 13th European conference, zurich, Switzerland, September 6-12, 2014, proceedings, part v 13},
  pages={740--755},
  year={2014},
  organization={Springer}
}

@inproceedings{dave2020tao,
  title={Tao: A large-scale benchmark for tracking any object},
  author={Dave, Achal and Khurana, Tarasha and Tokmakov, Pavel and Schmid, Cordelia and Ramanan, Deva},
  booktitle={Computer Vision--ECCV 2020: 16th European Conference, Glasgow, UK, August 23--28, 2020, Proceedings, Part V 16},
  pages={436--454},
  year={2020},
  organization={Springer}
}

@InProceedings{Chen_2019_CVPR,
author = {Chen, Zhao-Min and Wei, Xiu-Shen and Wang, Peng and Guo, Yanwen},
title = {Multi-Label Image Recognition With Graph Convolutional Networks},
booktitle = {Proceedings of the IEEE/CVF Conference on Computer Vision and Pattern Recognition (CVPR)},
month = {June},
year = {2019}
}

@InProceedings{Zhao_2021_ICCV,
    author    = {Zhao, Jiawei and Yan, Ke and Zhao, Yifan and Guo, Xiaowei and Huang, Feiyue and Li, Jia},
    title     = {Transformer-Based Dual Relation Graph for Multi-Label Image Recognition},
    booktitle = {Proceedings of the IEEE/CVF International Conference on Computer Vision (ICCV)},
    month     = {October},
    year      = {2021},
    pages     = {163-172}
}

@InProceedings{Xia_2023_ICCV,
    author    = {Xia, Xiaobo and Deng, Jiankang and Bao, Wei and Du, Yuxuan and Han, Bo and Shan, Shiguang and Liu, Tongliang},
    title     = {Holistic Label Correction for Noisy Multi-Label Classification},
    booktitle = {Proceedings of the IEEE/CVF International Conference on Computer Vision (ICCV)},
    month     = {October},
    year      = {2023},
    pages     = {1483-1493}
}

@incollection{keahey2020lessons,
  title={Lessons Learned from the Chameleon Testbed},
  author={Kate Keahey and Jason Anderson and Zhuo Zhen and Pierre Riteau and Paul Ruth and Dan Stanzione and Mert Cevik and Jacob Colleran and Haryadi S. Gunawi and Cody Hammock and Joe Mambretti and Alexander Barnes and Fran\c{c}ois Halbach and Alex Rocha and Joe Stubbs},
  booktitle={Proceedings of the 2020 USENIX Annual Technical Conference (USENIX ATC '20)},
  publisher={USENIX Association},
  month={July},
  year={2020}
}
}

\end{document}